\documentclass[10pt, a4paper]{article}

\usepackage{lrec-coling2024_new} 
\usepackage{verbatim}
\usepackage{booktabs}
\usepackage{amsfonts}
\usepackage{mathrsfs,amssymb}
\usepackage{amsthm,amsmath}
\usepackage{verbatim}
\usepackage{multirow}
\usepackage{xcolor}

\title{DELAN: Dual-Level Alignment for Vision-and-Language Navigation by Cross-Modal Contrastive Learning}
\name{
Mengfei Du\textsuperscript{${1\dag}$},
~Binhao Wu\textsuperscript{${1\dag}$},
~Jiwen Zhang\textsuperscript{$1$},
~Zhihao Fan\textsuperscript{$2$},\\
\large\bf 
~Zejun Li\textsuperscript{$1$},
~Ruipu Luo\textsuperscript{$1$},
~Xuanjing Huang\textsuperscript{$3$},
~Zhongyu Wei\textsuperscript{$1\ast$}
\thanks{$^{\ast}$ Corresponding author.}
\thanks{$^{\dag}$ Equal contribution.}
} 
\address{
\textsuperscript{1}School of Data Science, Fudan University, China\\
\textsuperscript{2}Alibaba Group, China\\
\textsuperscript{3}School of Computer Science, Fudan University, China\\
\{mfdu22, bhwu22, jiwenzhang21, rpluo21\}@m.fudan.edu.cn\\
\{zejunli20, xjhuang, zywei\}@fudan.edu.cn\\
}

\abstract{
Vision-and-Language navigation (VLN) requires an agent to navigate in unseen environment by following natural language instruction. For task completion, the agent needs to align and integrate various navigation modalities, including instruction, observation and navigation history. Existing works primarily concentrate on cross-modal attention at the fusion stage to achieve this objective. Nevertheless, modality features generated by disparate uni-encoders reside in their own spaces, leading to a decline in the quality of cross-modal fusion and decision.
To address this problem, we propose a \textbf{D}ual-lev\textbf{EL} \textbf{A}lig\textbf{N}ment (DELAN) framework by cross-modal contrastive learning. This framework is designed to align various navigation-related modalities before fusion, thereby enhancing cross-modal interaction and action decision-making. 
Specifically, we divide the pre-fusion alignment into dual levels: instruction-history level and landmark-observation level according to their semantic correlations. We also reconstruct a dual-level instruction for adaptation to the dual-level alignment. 
As the training signals for pre-fusion alignment are extremely limited, self-supervised contrastive learning strategies are employed to enforce the matching between different modalities.
Our approach seamlessly integrates with the majority of existing models, resulting in improved navigation performance on various VLN benchmarks, including R2R, R4R, RxR and CVDN.
\\ \newline 
}

\begin{document}

\maketitleabstract

\section{Introduction}
\begin{figure}[t]
  \setlength{\abovecaptionskip}{0.0cm}
  \centering
  \includegraphics[width=\linewidth]{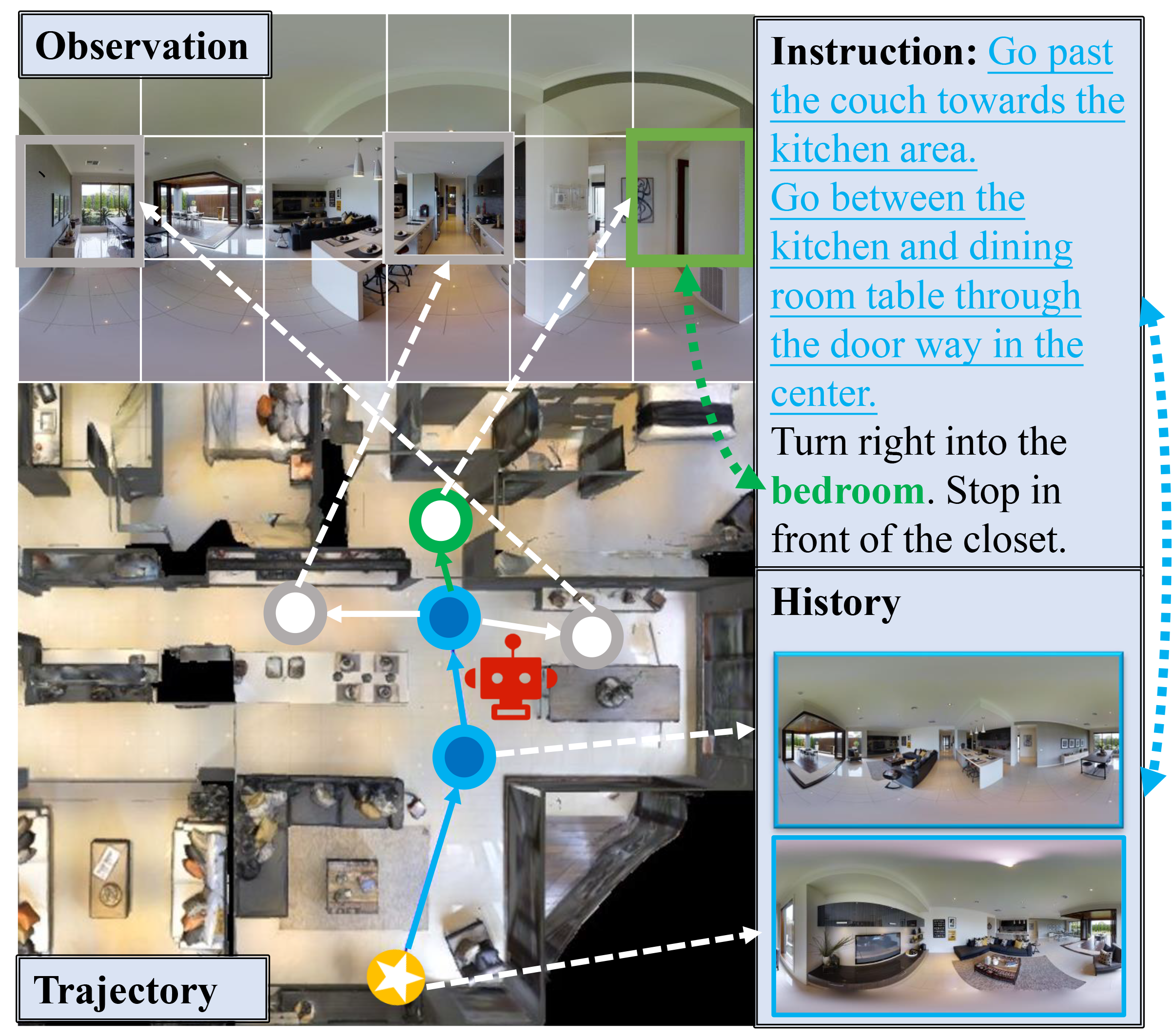}
  \caption[VLN process]{An Example. 
  The VLN agent needs to align the historical trajectory to the corresponding instructions (e.g. "Go past the couch towards\ldots") during navigation. At each time step, it selects one of the candidate viewpoints by evaluating the current observation in different directions and landmarks from the instructions (e.g. "bedroom").}
  \label{fig:example}
\end{figure}

Vision-and-Language navigation (VLN) has attracted increasing interest from natural language processing, computer vision and robotics communities due to its potential of applications such as embodied robots~\citep{gu2022vision}. The VLN task requires the agent to navigate to a predetermined destination, guided by natural language instruction and real-time visual observations. As depicted in Figure~\ref{fig:example}, a comprehensive navigation instruction typically comprises a sequence of action, direction and landmark clues. The agent needs to proficiently model its historical experiences and accurately relate the landmark clues to the environmental observations in order to successfully navigate.

\begin{figure}[t]
  \centering
  \includegraphics[width=\linewidth]{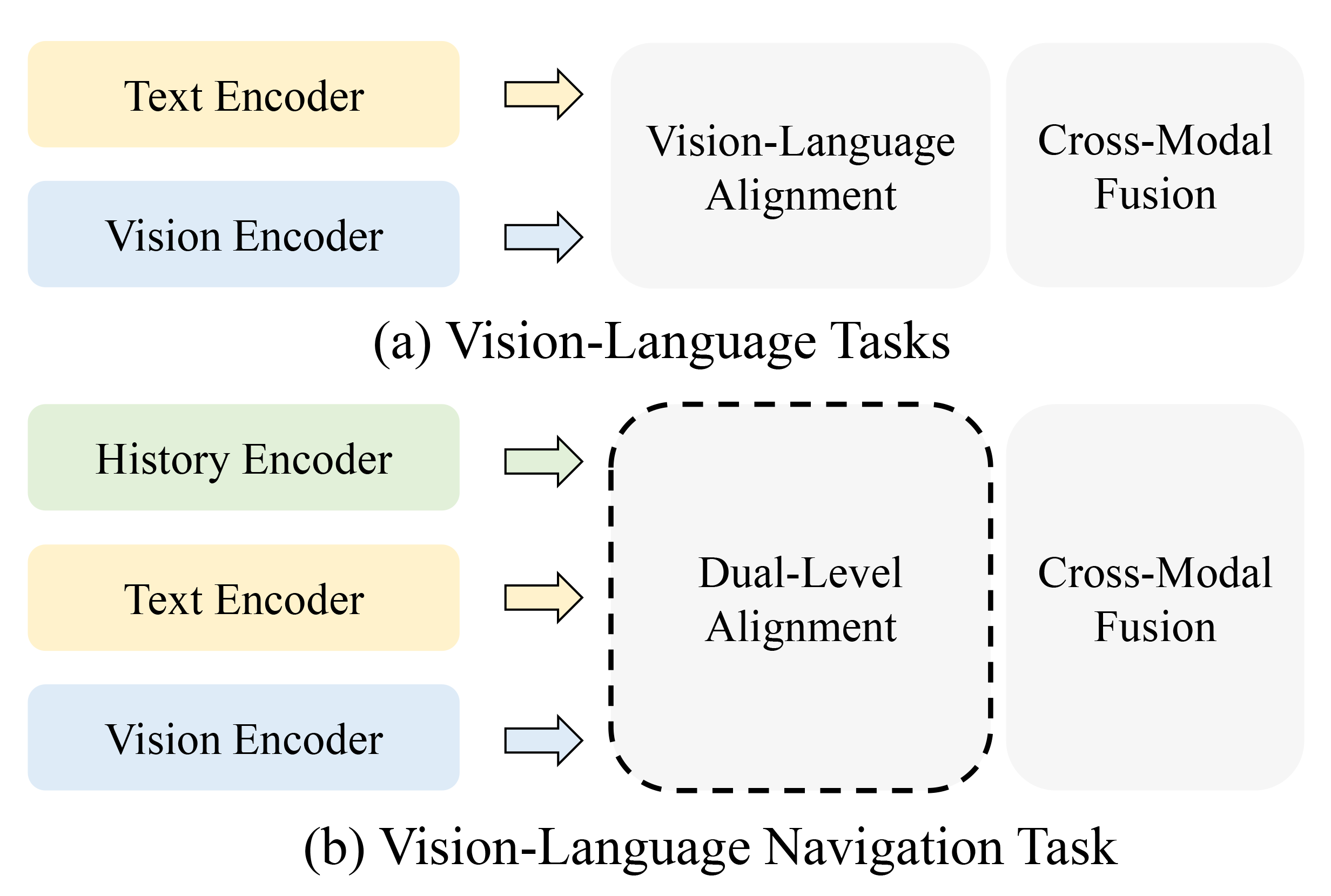}
  \caption[VLN process]{Comparison of pipelines for traditional VL and VLN tasks. (a) In VL tasks, vision-language pre-fusion alignment is often employed to encourage modality interaction. (b) However in VLN filed, there is a blank in pre-fusion alignment, which is solved by our dual-level alignment framework.}
  \label{fig:comparison}
\end{figure}

Similar to traditional Vision-Language task, the general model architecture of VLN agent adopts a multi-modal late-fusion paradigm~\citep{du2023uni}. Under this paradigm, individual modalities are encoded by their respective encoders and then passed through a fusion module, as illustrated in Figure~\ref{fig:comparison}. In Vision-Language task, aligning unimodal representations before fusion has been extensively validated as it contributes to cross-modal modeling ability in the fusion stage and semantic comprehension ability of unimodal encoders~\citep{li2021align, li2022mvptr}. However, pre-fusion alignment has not yet been explored in the context of VLN tasks, resulting in non-negligible disparities among correlated modalities before fusion.

Therefore, we propose to align corresponding unimodal representations and bridge the modality gaps prior to the cross-modal fusion stage in VLN task. This idea is challenging for two reasons. Firstly, there are limited training signals available for aligning unimodal representations. Training signals about navigation, such as navigation progress and relative spatial position, are used for supervising post-fusion representations in previous works~\citep{ma2019self,chen2021history}. Secondly, VLN is a sequential decision process distinct from traditional Vision-Language tasks~\citep{hong2021vln}.The navigation history condenses visited panorama and performed actions to retain past information, while observation at each time step contains detailed landmark semantics corresponding to the instruction. Consequently, how to effectively leverage the different characteristics of navigation history and real-time observations to enhance the cross-modal alignment remains an open question.

To address these problems, we propose a model-agnostic \textbf{D}ual-lev\textbf{EL} \textbf{A}lig\textbf{N}ment (DELAN) framework that utilizes cross-modal contrastive learning. As the training signals are limited, we propose the self-supervised contrastive learning approach, relying on the contrast between positive and negative pairs within a batch as the training signal. For the diverse characteristics of unimodal representations, we partition the cross-modal pre-fusion alignment into two separate levels: instruction-history level alignment and landmark-observation level alignment, according to their strong semantic correlations. Specifically, history modality records visual experiences along the described navigation trajectory. The landmarks, such as ``couch'' and ``closet'', appearing in the real-time observations constitute the textual foundation of language instructions. Furthermore, we restructure the original instruction into a dual-level instruction-landmark format to explicitly formulate the dual-level semantic correspondences with history and observation. For instruction-history alignment, we employ a contrastive loss on the history tokens and the instruction part of dual-level instruction across both global and local representations. For landmark-observation alignment, we use a contrastive loss on the observations and the landmark part of dual-level instruction at each time step across only local representations. 

We summarize our contributions as follows: (1) We are the first to introduce a cross-modal contrastive learning framework to improve the pre-fusion alignment in VLN task; (2) Inspired by the characteristics of different VLN modalities, we propose to conduct the dual-level alignment before fusion, i.e. instruction-history level alignment and landmark-observation level alignment, and develop self-supervised learning strategies respectively. (3) We validate our framework across various VLN benchmarks, demonstrating the effectiveness and consistency of our DELAN framework.\footnote{Code is available at \url{https://github.com/mengfeidu/DELAN}}

\section{Related Work}
\subsection{Vision-and-Language Navigation}
In recent years, training an instruction-following navigator has attracted growing research interest.
Earlier studies mainly employ a sequence-to-sequence LSTM framework in VLN tasks~\cite{fried2018speaker,tan2019learning}.
Owing to the success of transformer~\cite{Vaswani2017AttentionIA} in Vison-Language tasks~\citep{Tan2019LXMERTLC,lu2019vilbert,li2020oscar}, transformer-based models are widespread in the VLN field.
VLN$\circlearrowright$BERT~\cite{hong2021vln} devises a recurrent unit in the cross-modal transformer to model the navigation history. 
HAMT~\cite{chen2021history} proposes to encode all past observations and actions with hierarchical transformer.
DUET~\citep{chen2022think} utilizes a graph transformer to encode global history for efficient planning.
Besides the improvement of model architecture, various other methods are proposed to enhance the navigation performance, including training strategies~\cite{wang2019reinforced,Zhang2021CurriculumLF,liang2022contrastive} and auxiliary tasks~\cite{ma2019regretful,zhu2020vision}.

Despite the progress above, aligning navigation modalities remains a formidable challenge.
OAAM~\cite{qi2020object} and RelGraph~\cite{hong2020language} utilize separate soft-attention modules to implicitly align visual representations with different attribute words.
FGR2R~\cite{hong2020sub} grounds the observation to a certain sub-instruction using human annotations.
DDL~\citep{cheng2022learning} provides token-level supervision by labeling landmark and action words relating to observation at each time step.
However, these approaches only focus on the cross-modal fusion module, neglecting the modality gaps among unimodal representations that are commonly appeared in vision-language tasks~\cite{li2021align}. 
In contrast, our work presents a self-supervised framework to align these modalities before fusion stage by contrastive learning to promote cross-modal reasoning.

\begin{figure*}[htbp]
  \centering
  \includegraphics[width=\textwidth]{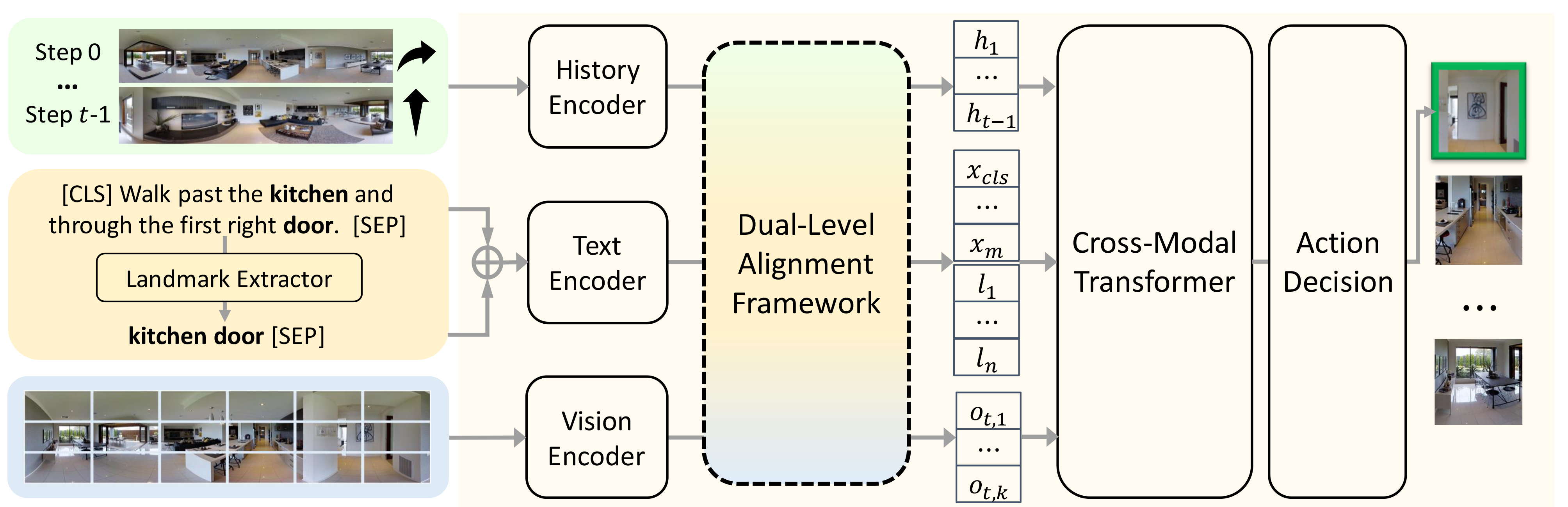}
  \caption{The pipeline of VLN model enhanced with DELAN framework. Landmark words are extracted from the original instructions, and three unimodal encoders encode four types of tokens: instructions, landmarks, histories, and observations.  Then, the alignments are employed at pre-fusion stage to enhance cross-modal interaction and action decision.}
  \label{fig:overview_split1}
\end{figure*}
\subsection{Cross-modal Contrastive Learning}
Cross-modal contrastive learning has proven its effectiveness in modeling vision-language relationships and has gained prominence as a popular strategy in VL tasks. CLIP~\cite{radford2021learning} firstly employs cross-modal contrastive learning on large-scale image-text pairs. FILIP~\cite{yao2021filip} utilizes token-wise maximum similarity to achieve the fine-grained alignment between visual and textual tokens. TACo~\cite{yang2021taco} proposes a token-aware contrastive loss and a cascade sampling method for aligning video and text. X-CLIP~\cite{ma2022x} proposes attention method to realize multi-grained contrast between text and video. 
CITL~\cite{liang2022contrastive} first introduces the contrastive learning into VLN domain but only considers the unimodal contrast.
Inspired by these works, we propose to apply contrastive learning to bridge the modality gaps. Furthermore, to handle the sparse training signals for pre-fusion alignment, we apply self-supervised learning strategies to realize our dual-level alignment framework.


\section{Preliminaries}
\subsection{Problem Formulation} 
Given the instruction $\mathcal{I}=\{w_1,w_2,...,w_m\}$, the agent receives a panorama observation $\mathcal{O}_t$ of its surrounding environment at each step $t$. The panorama observation consists of $k$ single view images with corresponding relative angles, i.e. $\mathcal{O}_t\triangleq([v_1^o;a_1^o],...,[v_k^o,a_k^o])$, where $v_i^o$ is the visual feature of the $i$-th view and $a_i^o$ is the relative angle with respect to the current view (subscript $t$ is omitted for simplicity). At each viewpoint, the agent has $n_t$ navigable candidate viewpoints $\mathcal{O}^c_t\triangleq([v_1^c;a_1^c],...,[v_{n_t}^c;a_{n_t}^c])$. All observations $\mathcal{O}_i$ and performed actions $a_i^h$ before time step $t$ form the history $\mathcal{H}_t\triangleq([\mathcal{O}_1;a_1^h],...,[\mathcal{O}_{t-1};a_{t-1}^h])$, where action $a_i^h$ denotes the turned angles at step $i$. The navigation is finished when the agent makes \textbf{STOP} decision. The task is considered accomplished when the agent stops near the target location ($<$3m).

\subsection{Base Agent Model}
Current VLN agents~\cite{chen2021history,chen2022think} mainly adopts multi-modal late-fusion learning paradigm~\cite{du2023uni}. To encode each modality, they usually employ standard BERT~\cite{devlin2018bert} for textual instructions and vision transformer (ViT)~\cite{dosovitskiy2020image} for visual observations. As for history modality, they encode historical observations and performed actions during navigation, enabling agents to remember the trajectory. Specially, graph-based methods~\cite{chen2022think} maintain a topological map to record the historical experiences.
Above three unimodal encoding processes can be formulated as follows:
\begin{equation}
\label{eq_unimodal}
\begin{aligned}
    \{x_{cls},x_1,x_2,\dots,x_m\}&=\mathrm{TextEnc}([[CLS];\mathcal{I}])\\
    \{o_{t,1},o_{t,2},\dots,o_{t,k}\}&=\mathrm{VisionEnc}(\mathcal{O}_t)\\
    \{h_1,h_2,\dots,h_{t-1}\}&=\mathrm{HistoryEnc}(\mathcal{H}_t)
\end{aligned}
\end{equation}
where $x_i,o_{t,i},h_i \in \mathbb{R}^d$. These unimodal embeddings are further fed to a cross-modal transformer to predict the action probability over the candidate viewpoints $\mathcal{O}^c_t$. The candidate viewpoint with largest probability will be chosen as next move location.

\subsection{Contrastive Learning Base}
\label{3_3}
\paragraph{Similarity reduce function.}
Given a similarity vector/matrix $M \in \mathbb{R}^{n\times m}$ of two instances from distinct modalities, we need to reduce it to scalar value for later contrastive loss calculation. Attention mechanism is introduced in the aggregation process, allowing each element to attend to multiple contrasting elements and automatically balance their influences~\citep{ma2022x}. Concretely, we first perform attention operations on row and column to reduce values in each dimension.
\begin{equation}
\begin{aligned}
    M_r=\sum_{i=1}^n\frac{\exp(M_{(i,*)})}{\sum_{j=1}^n\exp(M_{(j,*)})}\odot M_{(i,*)}\\
    M_c=\sum_{i=1}^m\frac{\exp(M_{(*,i)})}{\sum_{j=1}^m\exp(M_{(*,j)})}\odot M_{(*,i)}
\end{aligned}
\end{equation}
where $*$ represents all elements in the dimension, $\odot$ represents element-wise product, $M_r \in \mathbb{R}^{1\times m}$ and $M_c \in \mathbb{R}^{n\times 1}$.
Likewise, we second perform attention operations on column and row to get a scale value respectively.
\begin{equation}
\begin{aligned}
    M_r^{\prime}=\sum_{i=1}^m\frac{\exp({M_r}_{(1,i)})}
    {\sum_{j=1}^m\exp({M_r}_{(1,j)})}{M_r}_{(1,i)}\\
    M_c^{\prime}=\sum_{i=1}^n\frac{\exp({M_c}_{(i,1)})}
    {\sum_{j=1}^n\exp({M_c}_{(j,1)})}{M_c}_{(i,1)}
\end{aligned}
\end{equation}
where $M_r^{\prime}, M_c^{\prime} \in \mathbb{R}^1$.
Finally, we average two scale values to acquire similarity score of two instances and define above process as reduce function $\mathrm{R}$.
\begin{equation}
\label{eq_reduce_function}
    \mathrm{R}(M)=(M_r^{\prime}+M_c^{\prime})/2
\end{equation}

\paragraph{General contrastive loss.}
Under a general formulation of cross-modal contrastive learning~\cite{radford2021learning}, we consider $B$ pairs $\{\mathbf{t}^{(i)},\mathbf{v}^{(i)}\}_{i=1}^B$ from two modalities $\mathbf{t}$ and $\mathbf{v}$ in each training batch. 
With similarity scores $\{sim(\mathbf{t}^{(i)},\mathbf{v}^{(j)})\}_{i,j=1}^B \in \mathbb{R}$ among instances $\mathbf{t}^{(i)}$ and $\mathbf{v}^{(j)}$, we can compute both $\mathbf{t}$-to-$\mathbf{v}$ and $\mathbf{v}$-to-$\mathbf{t}$ contrastive losses.
The total contrastive loss can be summed from above losses and defined as contrastive loss function $\mathrm{L}$ as:

\begin{gather}
    {loss}(\mathbf{t},\mathbf{v})=-\frac{1}{B}\sum_{i=1}^{B}
    \log \frac{\exp(sim(\mathbf{t}^{(i)},\mathbf{v}^{(i)})/\tau)}{\sum_{j=1}^B\exp(sim(\mathbf{t}^{(i)},\mathbf{v}^{(j)})/\tau)} \nonumber \\
    {loss}(\mathbf{v},\mathbf{t})=-\frac{1}{B}\sum_{i=1}^{B}
    \log \frac{\exp(sim(\mathbf{v}^{(i)},\mathbf{t}^{(i)})/\tau)}
    {\sum_{j=1}^B\exp(sim(\mathbf{v}^{(i)},\mathbf{t}^{(j)})/\tau)} \nonumber \\
    \mathrm{L}(\mathbf{t},\mathbf{v})={loss}(\mathbf{t},\mathbf{v})+{loss}(\mathbf{v},\mathbf{t})
\label{eq_universal_cl}
\end{gather}
where $\tau$ is temperature parameter.

\section{Methods}
\begin{figure*}[htbp]
  \centering
  \includegraphics[width=\textwidth]{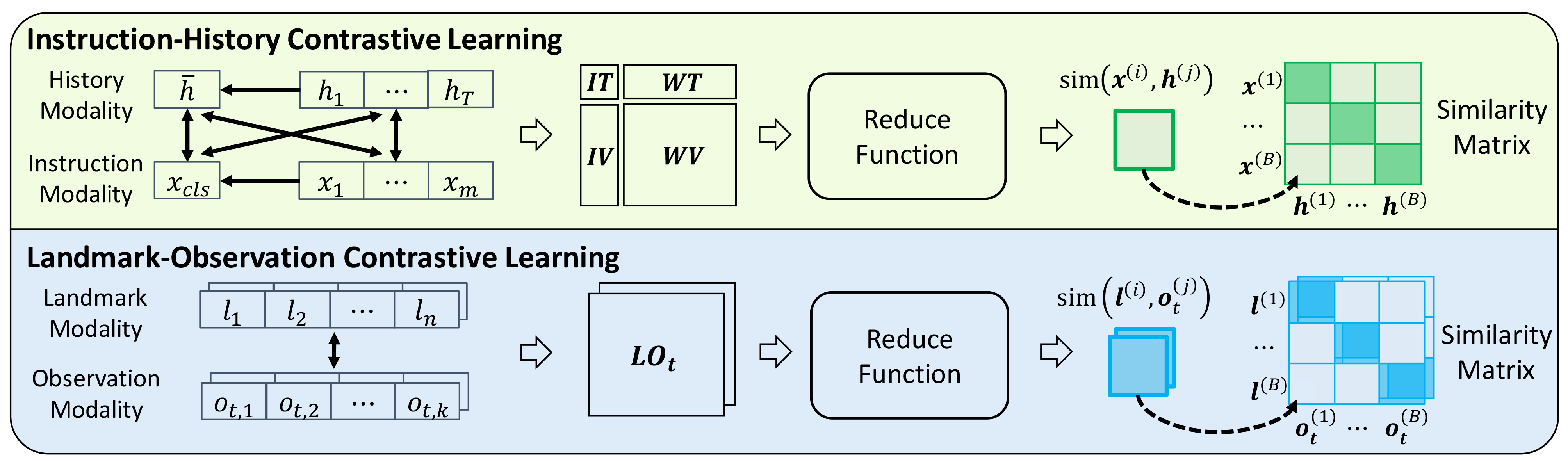}
  \caption{Overview of our DELAN framework. The alignments are employed at both instruction-trajectory and landmark-observation levels. Specifically, we implement instruction-trajectory contrast across global (\textbf{I}nstruction, \textbf{T}rajectory) and local (\textbf{W}ord, \textbf{V}iewpoint) representations using complete instructions and trajectories. At the landmark-observation level, we perform contrast at each time step. }
  \label{fig:overview_split2}
\end{figure*}

In this section, we elaborate each component of our dual-level alignment (DELAN) framework. We explain the dual-level instruction construction in section \ref{4_1}. Then we introduce the instruction-history level alignment in section~\ref{4_2}, and the landmark-observation level alignment in section~\ref{4_3}, both by self-supervised contrastive learning. Finally, we present the complete training strategy in section~\ref{4_4}. The pipeline of our framework is illustrated as Figure~\ref{fig:overview_split1} and Figure~\ref{fig:overview_split2}.

\subsection{Dual-level Instruction Construction}
\label{4_1}
A navigation instruction (e.g. "Walk to the right past the TV.") comprises a series of action, orientation and landmark words. 
Since the navigation history has strong semantic correlation with the entire instruction, while the real-time observations are closely related to the landmark words in instructions, it is significant to explicitly represent such dual-level semantic correspondences.
Therefore, an additional landmark part is introduced for alignment with observation while the original instruction part is reserved for alignment with history.

Specifically, we concatenate original instruction with the extracted landmark words to form a dual-level instruction. We employ a standard pre-trained BERT model~\cite{vaswani2017attention} fine-tuned on part-of-speech task to extract all nouns in the instruction as landmark words.
The instruction level text guides the overall navigation and the landmark level text helps focus on key details in observation. After this we can drive the instruction part align to history modality and the landmark part align to observation modality respectively.

Formally, we extract $n$ landmark words $\mathcal{I}^l=\{w_1^l,w_2^l,\dots,w_n^l\}$ from original instructions. These landmark words are appended to the original instructions, subsequently fed into the text encoder to get textual embeddings, i.e.  
\begin{equation}
\begin{split}
\label{eq_tpinstr}
     \{x_{cls},X, L\}=\mathrm{TextEnc}([[\text{CLS}];\mathcal{I};\mathcal{I}^l])
\end{split}
\end{equation}
where $X=(x_1,x_2,\dots,x_m)$ is the instruction word matrix, $L=(l_1,l_2,\dots,l_n)$ is the landmark word matrix. The representation of navigation history and real-time observations are computed based on equation~\ref{eq_unimodal}, resulting in history embeddings matrix $H=(h_1,h_2,\dots,h_T)$ and observation embeddings matrix $O_t=(o_{t,1},o_{t,2},\dots,o_{t,k})$ at time step $t$. Note that $T$ is the total number of navigation steps.

\subsection{Instruction-History Alignment}
\label{4_2}
In VLN task, it's crucial to model historical experiences which helps agent make informed decision based on past information. The history modality exhibits a strong semantic correlation with instruction. The instruction comprehensively describes the navigation trajectory while the history records the experience along this trajectory. The instruction token sequence and history token sequence both adhere to chronological order and are synchronized in time. Current unimodal representation of history is solely derived from past observation and action~\cite{chen2021history,cheng2022learning} but lack adequate supervision from textual contents before the fusion stage. 
Hence, we propose to align the history modality with the instruction part of dual-level instruction. Since they both represent the trajectory and are synchronized in time, exhibiting an element-wise correspondence, we implement the contrastive loss across global representation and local representation.
Concretely, we split the instruction-history level alignment into instruction-trajectory alignment, word-trajectory alignment, instruction-viewpoint alignment and word-viewpoint alignment, each providing different benefits.

Additionally, we introduce the global trajectory representation $\bar{h}$ which obtained through mean pooling of all history embeddings during navigation and choose the textual $\text{[CLS]}$ embedding $x_{cls}$ as the global instruction representation.

\begin{table*}[htbp]
    \setlength{\abovecaptionskip}{0.15cm}
    \setlength{\belowcaptionskip}{0cm}
    \centering

    \resizebox{1\textwidth}{!}{
    \begin{tabular}{l|cccc|cccc|cccc}
    \toprule
    \multirow{2}{*}{Model}&\multicolumn{4}{c}{Val-Seen}
    &\multicolumn{4}{|c}{Val-Unseen}&\multicolumn{4}{|c}{Test-Unseen} \\
    \cmidrule(lr){2-13}
    & TL & NE$\downarrow$ & SR$\uparrow$ & SPL$\uparrow$ &
      TL & NE$\downarrow$ & SR$\uparrow$ & SPL$\uparrow$ &
      TL & NE$\downarrow$ & SR$\uparrow$ & SPL$\uparrow$ \\
    \cmidrule(lr){1-13}
    SF~\citeyearpar{fried2018speaker}  & 
    - & 3.36 & 66 & - &
    - & 6.62 & 35 & - &
    14.82 & 6.62 & 35 & 28 \\
    
    EnvDrop~\citeyearpar{wang2020environment}  & 
    11.00 & 3.99 & 62 & 59 &
    10.70 & 5.22 & 52 & 48 &
    11.66 & 5.23 & 51 & 47 \\

    OAAM~\citeyearpar{qi2020object}  &
    10.20 & - & 65 & 62 &
    9.95 & - & 54 & 50 &
    10.40 & - & 53 & 50 \\

    PREVALENT~\citeyearpar{hao2020towards}  &
    10.32 & 3.67 & 69 & 65 &
    10.19 & 4.71 & 58 & 53 &
    10.51 & 5.30 & 54 & 51 \\

    RelGraph~\citeyearpar{hong2020language}  & 
    10.13 & 3.47 & 67 & 65 &
    9.99 & 4.73 & 57 & 53 &
    10.29 & 4.75 & 55 & 52 \\
    
    SSM~\citeyearpar{wang2021structured}  &
    14.70 & 3.10 & 71 & 62 &
    20.70 & 4.32 & 62 & 45 &
    20.40 & 4.57 & 61 & 46 \\
    
    VLN$\circlearrowright$BERT~\citeyearpar{hong2021vln}  & 
    11.13 & 2.90 & 72 & 68 &
    12.01 & 3.93 & 63 & 57 &
    12.35 & 4.09 & 63 & 57 \\

    LOViS~\citeyearpar{zhang2022lovis}  &
    - & 2.40 & 77 & 72 &
    - & 3.71 & 65 & 59 &
    - & 4.07 & 63 & 58 \\
    
    DUET~\citeyearpar{chen2022think} &
    12.32 & 2.28 & 79 & 73 & 
    13.94 & 3.31 & 72 & 60 &
    14.74 & 3.65 & 69 & 59 \\

    \cmidrule(lr){1-13}
    HAMT~\citeyearpar{chen2021history}  &
    11.15 & 2.51 & 75.61 & 72.18 &
    11.46 & 3.62 & 66.24 & 61.51 &
    12.27 & 3.93 & 65.09 & 60.02 \\
    HAMT+DELAN & 
    11.16 & 2.42 & 76.89 & 73.61 & 
    11.62 & 3.46 & 67.22 & 61.53 & 
    12.34 & 3.96 & 65.30 & 60.70 \\
    \cmidrule(lr){1-13}
   $\text{DUET}^{*}$~\citeyearpar{chen2022think} &
    14.32 & 2.38 & 76.69 & 69.74 &
    14.52 & 2.95 & 72.80 & 61.70 &
    13.99 & 3.27 & 71.07 & 60.99 \\
    DUET+DELAN & 
    12.02 &\textbf{1.94} & \textbf{81.19} & \textbf{76.66} & 
    13.07 & \textbf{2.94} & \textbf{73.35} & \textbf{63.39} & 
    12.98 & \textbf{3.27} & \textbf{71.22} & \textbf{62.69} \\
    \bottomrule
    \end{tabular}}
    \caption{Comparison of performance with SOTA methods on R2R dataset.}
    \label{tab:main_result_r2r}
\end{table*}

\paragraph{Instruction-Trajectory Alignment}
helps global alignment. Given global instruction representation $x_{cls}\in \mathbb{R}^{d}$ and global trajectory representation $\bar{h}\in \mathbb{R}^{d}$, the similarity between instruction and trajectory can be computed by matrix multiplication as:
\begin{equation}
    S_{IT}=\bar{h}^{\top}x_{cls},
\end{equation}
where $S_{IT}\in \mathbb{R}^{1}$ represents for the instruction-trajectory similarity score.

\paragraph{Word-Trajectory Alignment}
assists in emphasizing the most important words and downplaying words with low navigation information, like prepositions and conjunctions. For given local word representation $X\in \mathbb{R}^{m\times d}$ and global trajectory representation $\bar{h}\in \mathbb{R}^{d}$, we also use matrix multiplication to get the similarity between word and trajectory:
\begin{equation}
    S_{WT}=(X\bar{h})^{\top},
\end{equation}
where $S_{WT}\in \mathbb{R}^{1\times m}$ represents for the word-trajectory similarity score vector.

\paragraph{Instruction-Viewpoint Alignment}
facilitates generalizing viewpoints which closely follow the instruction. We can compute the similarity between instruction and viewpoint by global instruction representation $x_{cls}\in \mathbb{R}^{d}$ and local viewpoint representation (equal to history representation) $H\in \mathbb{R}^{T\times d}$, which can be formulated as follows:
\begin{equation}
    S_{IV}=Hx_{cls},
\end{equation}
where $S_{IV}\in \mathbb{R}^{T\times 1}$ represents for the instruction-viewpoint similarity score vector.

\paragraph{Word-Viewpoint Alignment}
contributes to monitoring the navigation progress, as each viewpoint embedding can implicitly align with its corresponding sub-instruction word embeddings. The similarity between word and viewpoint can also be obtained from local word representation $X\in \mathbb{R}^{m\times d}$ and local viewpoint representation $H\in \mathbb{R}^{T\times d}$ as:
\begin{equation}
    S_{WV}=HX^{\top},
\end{equation}
where $S_{WV}\in \mathbb{R}^{T\times m}$ represents for word-viewpoint similarity score matrix.
We employ the reduce function $\mathrm{R}$ in equation~\ref{eq_reduce_function} to derive scalar values from similarity score vectors or matrices, based on attention mechanisms. Given $B$ instruction-history pairs $\{\mathbf{x}^{(i)},\mathbf{h}^{(i)}\}_{i=1}^B$ where $\mathbf{x}^{(i)}=\{x_{cls}^{(i)},X^{(i)}\}$ represents in-batch $i$-th instruction instance and $\mathbf{h}^{(i)}=\{{\bar{h}}^{(i)},H^{(i)}\}$ represents in-batch $i$-th history instance, the score $sim(\mathbf{x}^{(i)},\mathbf{h}^{(j)})$ measures the semantic similarity between two instances and can be averaged from various similarity scores:
\begin{equation}
\begin{split}
    sim(\mathbf{x}^{(i)},\mathbf{h}^{(j)})&=(S_{IT}+\mathrm{R}(S_{WT}) \\
    &+\mathrm{R}(S_{IV})+\mathrm{R}(S_{WV}))/4
\end{split}
\label{eq_s_ih}
\end{equation}
Finally, the instruction-history level contrastive loss can be calculated from equation~\ref{eq_universal_cl} and equation~\ref{eq_s_ih}:
\begin{equation}
    \mathcal{L}_{IH}=\mathrm{L}(\mathbf{x},\mathbf{h})
\end{equation}

\subsection{Landmark-Observation Alignment}
\label{4_3}
The panoramic observation at each time step consists of $k$ view images together with relative directions, providing a detailed spatial layout.
Such observation exhibits strong semantic correlation with landmarks in instructions, as the navigation can be typically succeeded by following landmark entities in the environment based on landmark references in instructions. The pre-fusion alignment between landmark and observation enhance the interaction between related representations.

Consequently, we align the observation modality with the landmark part of dual-level instructions. The observation and landmarks exhibit an element-wise correspondence, as landmarks only present in some of the view images and real-time observation contains only a subset of the landmarks mentioned in instruction. Therefore, we implement the contrastive loss only across local representation here.

\paragraph{Landmark-Observation Alignment}
helps align landmark tokens with specific view images of observation at each time step. The similarity matrix can be obtained from local landmark representation $L\in \mathbb{R}^{n\times d}$ and local representation observation $O_t\in \mathbb{R}^{k\times d}$ at time step $t$ as follows:
\begin{equation}
\label{eq_S_l_o}
    S_{LO_t}= O_t L^T,
\end{equation}
where $S_{LO_t}\in \mathbb{R}^{k\times n}$ is the landmark-observation similarity score matrix at step $t$.
We also employ the reduce function $\mathrm{R}$ in equation~\ref{eq_reduce_function} to dynamically score the importance of each element in similarity matrix and aggregate them.
Analogously, given $B$ landmark-observation pairs $\{\mathbf{l}^{(i)},\mathbf{o}_{t}^{(i)}\}_{i=1}^B$ where $\mathbf{l}^{(i)}=\{L^{(i)}\}$ represents in-batch $i$-th landmark instance and $\mathbf{o}_{t}^{(i)}=\{O_t^{(i)}\}$ represents in-batch $i$-th observation instance at time step $t$, the similarity score $sim(\mathbf{l}^{(i)},\mathbf{o}_t^{(j)})$ measures the semantic similarity between the landmark and observation :
\begin{equation}
\begin{split}
    sim(\mathbf{l}^{(i)},\mathbf{o}_t^{(j)})=\mathrm{R}(S_{LO_t})
\end{split}
\label{eq_s_lo}
\end{equation}
Finally, the landmark-observation level contrastive loss can be computed from averaged similarity scores along the trajectory according to equation~\ref{eq_universal_cl} and equation~\ref{eq_s_lo}. The formula is: 
\begin{equation}
    \mathcal{L}_{LO}=\frac{1}{T}\sum_{t=1}^T\mathrm{L}(\mathbf{l},\mathbf{o}_t)
\end{equation}
Note that $T$ is the trajectory length.
\begin{table*}[htbp]
    \centering
    \setlength{\abovecaptionskip}{0.15cm}
    \setlength{\belowcaptionskip}{0cm}
    \resizebox{1\textwidth}{!}{
    \begin{tabular}{l|cccc|cccc|cccc}
    \toprule
    \multirow{2}{*}{Model}&\multicolumn{4}{c}{Val-Seen}
    &\multicolumn{4}{|c}{Val-Unseen}&\multicolumn{4}{|c}{Test-Unseen} \\
    \cmidrule(lr){2-13}
    & SR$\uparrow$ & SPL$\uparrow$ & nDTW$\uparrow$ & sDTW$\uparrow$ &
      SR$\uparrow$ & SPL$\uparrow$ & nDTW$\uparrow$ & sDTW$\uparrow$ &
      SR$\uparrow$ & SPL$\uparrow$ & nDTW$\uparrow$ & sDTW$\uparrow$ \\
    \cmidrule(lr){1-13}
    Multilingual Baseline~\citeyearpar{ku2020room} & 
    25.2 & - & 42.2 & 20.7 &
    22.8 & - & 38.9 & 18.2 &
    21.0 & 18.6 & 41.1 & 20.6 \\
    
    Monolingual Baseline~\citeyearpar{ku2020room} & 
    28.8 & - & 46.8 & 23.8 &
    28.5 & - & 44.5 & 23.1 &
    25.4 & 22.6 & 41.1 & 32.4 \\

    EnvDrop-CLIP~\citeyearpar{wang2020environment}  & 
    - & - & - & - &
    42.6 & - & 55.7 & - &
    38.3 & - & 51.1 & 32.4 \\
    
    \cmidrule(lr){1-13}
    Multilingual HAMT~\citeyearpar{chen2021history} &
    59.4 & \textbf{58.9} & 65.3 & 50.9 &
    56.5 & \textbf{56.0} & 63.1 & 48.3 &
    53.1 & 46.6 & 59.9 & 45.2 \\
    Multilingual HAMT+DELAN & 
    \textbf{62.1} & 58.5 &  \textbf{67.9} &  \textbf{53.4} & 
    \textbf{57.5} & 53.9 &  \textbf{65.4} &  \textbf{49.5} & 
    \textbf{53.9} & \textbf{47.7} &  \textbf{60.9} &  \textbf{45.9} \\

    \bottomrule
    \end{tabular}}
    \caption{Comparison of performance with SOTA methods on RxR dataset.}
    \label{tab:main_result_rxr}
    \vspace{-0.2cm}
\end{table*}

\begin{table}[t]
    \centering
    \setlength{\abovecaptionskip}{0.15cm}
    \setlength{\belowcaptionskip}{0cm}
    \resizebox{0.5\textwidth}{!}{
    \begin{tabular}{l|ccccc}
    \toprule
    \multirow{2}{*}{Model}&\multicolumn{5}{c}{Val-Unseen} \\
    \cmidrule(lr){2-6}
    & NE$\downarrow$ & SR $\uparrow$ & CLS$\uparrow$ & nDTW$\uparrow$ & sDTW$\uparrow$\\
    \cmidrule(lr){1-6}
    SF~\citeyearpar{fried2018speaker} & 
    8.47 & 24.0 & 30.0 & - & - \\
    
    RCM~\citeyearpar{wang2019reinforced}  & 
    8.08 & 26.0 & 35 & 30 & 13 \\
    
    
    OAAM~\citeyearpar{qi2020object} & 
    8.51 & 26.6 & 36.2 & 30.3 & 12.6 \\
    
    RelGraph~\citeyearpar{hong2020language} &
    7.43 & 36.0 & 41.0 & 47.0 & 34.0 \\ 
    
    VLN$\circlearrowright$BERT~\citeyearpar{hong2021vln} &
    6.48 & 43.6 & 51.4 & 45.1 & 29.9\\
    
    DDL~\citeyearpar{cheng2022learning}&
    6.43 & 42.4 & 43.6 & 38.5 & 21.0\\
    
    LOViS~\citeyearpar{zhang2022lovis}&
    6.07 & 45.0 & 45.0 & 43.0 & 23.0\\

    DUET~\citeyearpar{chen2022think}&
    5.60 & 51.2 & 47.0 & 42.7 & 29.3\\

    \cmidrule(lr){1-6}
    HAMT~\citeyearpar{chen2021history}  & 6.09 & 44.6 & 57.7 & 50.3 & 31.8\\
    HAMT+DELAN & 
    5.99 & 45.5  & \textbf{61.0} & \textbf{54.0} & \textbf{34.4} \\
    \cmidrule(lr){1-6}
    $\text{DUET}^{*}$~\citeyearpar{chen2022think} & \textbf{5.42} & 51.4 & 46.0 & 40.2 & 27.1\\
    DUET+DELAN & 5.47 & \textbf{52.5} & 47.6 & 41.6 & 28.9 \\
    \bottomrule
    \end{tabular}}
    \caption{Comparison of performance with SOTA methods on R4R dataset.  }
    \label{tab:main_result_r4r}

\end{table}

\subsection{Training Strategy}
\label{4_4}
Two typical training strategies in VLN task are Imitation Learning (IL) and Reinforcement Learning (RL). In IL phase, the agent learns by following teacher action $a_t^{*}$:
$\mathcal{L}_{IL}=-\frac{1}{T}\sum_{t=1}^{T}a_t^{*}\log(p_t)$,
where $p_t$ is predicted action probability. In RL phase, the agent samples the action to explore the environment and learns from the reward: 
$\mathcal{L}_{RL}=-\frac{1}{T}\sum_{t=1}^{T}a_t^s\log(p_t)A_t$, 
where $a_t^s$ is the sampled action according to predicted action probability $p_t$ and $A_t$ is the advantage in A2C algorithm~\cite{mnih2016asynchronous}.
Overall, by combining contrastive loss on instruction-history level and landmark-observation level, the total loss is calculated as:
\begin{equation}
\label{eq_total_loss}
    \mathcal{L}_{loss}=
    \lambda_1 \mathcal{L}_{RL}
    +\lambda_2 \mathcal{L}_{IL}
    +\lambda_3 \mathcal{L}_{IH}
    +\lambda_4 \mathcal{L}_{LO}
\end{equation}
where $\lambda_1$, $\lambda_2$, $\lambda_3$ and $\lambda_4$ are weighting parameters that balance different losses.

\section{Experiments}
\label{5}
\subsection{Experimental Setup}
\paragraph{Datasets.}
We evaluated our method on three representative VLN tasks (four datasets): VLN with fine-grained instructions R2R~\citep{anderson2018vision} and RxR~\citep{ku2020room}; long-horizon VLN R4R~\citep{jain2019stay}; and vision-and-dialogue navigation 
CVDN~\citep{thomason2020vision}.
\begin{itemize}
    \setlength{\itemsep}{0pt}
    \setlength{\parsep}{0pt}
    \setlength{\parskip}{0pt}
    \item \textbf{R2R} is the basic VLN dataset which contains step-by-step navigation instructions. It includes 90 scans with 22k human-annotated instruction-trajectory pairs. The dataset is divided into train, val seen, val unseen and test unseen splits,consisting of 61, 56, 11 and 18 scans respectively. Scans in the unseen splits are excluded from the training set.
    \item \textbf{RxR} is a multilingual (English, Hindi and Telugu) VLN dataset. Comparing to R2R, RxR is 10x larger with longer and more various paths.
    \item \textbf{R4R} extends R2R dataset by concatenating two adjacent trajectories and their instructions. It encourages the agent to navigate as human plan instead of going to the destination directly.
    \item \textbf{CVDN} requires an agent to navigate based on multi-turn question-answering dialogues. The lengths of instructions and trajectories are much longer than R2R. 
\end{itemize}

\begin{table}[t]
    \centering
    \setlength{\abovecaptionskip}{0.15cm}
    \setlength{\belowcaptionskip}{0cm}
    \resizebox{0.5\textwidth}{!}{
    \begin{tabular}{l|ccc}
    \toprule
    Model&Val-Seen &Val-Unseen&Test-Unseen \\
    \cmidrule(lr){1-4}
    PREVALENT~\citeyearpar{hao2020towards} & - & 3.15 & 2.44\\
    VISITRON~\citeyearpar{shrivastava2021visitron} & 5.11 & 3.25 & 3.11\\
    MT-RCM+EnvAg~\citeyearpar{wang2020environment} & 5.07 & 4.65 & 3.91\\
    MTVM~\citeyearpar{lin2022multimodal} & - & 5.15 & 4.82\\
    \cmidrule(lr){1-4}
    HAMT~\citeyearpar{chen2021history} & 6.91 & 5.13 & 5.58\\
    HAMT+DELAN & 7.18 & 5.20 & 5.85 \\
    \cmidrule(lr){1-4}
    $\text{DUET}^{*}$~\citeyearpar{chen2022think} & \textbf{8.57} & 5.58 & 5.96\\
    DUET+DELAN & 7.40 & \textbf{5.95} & \textbf{6.19} \\    
    \bottomrule
    \end{tabular}}
    \caption{Comparison of goal progress with SOTA methods on CVDN dataset.}
    \label{tab:main_result_cvdn}
\end{table}

\paragraph{Evaluation Metrics.}
For R2R, we use standard evaluation metrics following previous work~\cite{anderson2018evaluation,anderson2018vision}: Trajectory Length (TL), Navigation Error (NE), Success Rate (SR), Success rate weighted by Path Length (SPL).
Metrics that measure the path fidelity between the predicted path and target path are introduced for R4R and RxR, including normalized Dynamic Time Warping (nDTW), Success weighted by normalized Dynamic Time Warping (SDTW)~\citep{ilharco2019general} and Coverage weighted by Length Score (CLS)~\citep{jain2019stay}. 
For CVDN, we use Goal Progress (GP) in meters~\citep{thomason2020vision} as the primary metric. GP measures the difference between completed distance and left distance to the goal, so the higher the better. Note that \textbf{bold} metrics in tables indicate the best results.

\paragraph{Implementation Details.}
The experiments are conducted on a NVIDIA 3090 GPU.
For contrastive loss, we set $\lambda_3 = 0.01$ and $\lambda_4 = 0.1$ in equation ~\ref{eq_total_loss} while keeping temperature parameter $\tau = 1$. We also introduce the memory bank~\cite{wu2018unsupervised} to expand number of negative samples and fix the size to 480.
Our baseline models consist of HAMT~\citep{chen2021history} and DUET~\citep{chen2022think}.
Different from the original implementation~\citep{chen2022think}, we utilize CLIP-ViT-B-16~\citep{li2022envedit} as vision encoder for DUET. The batch size is set to 4 for DUET and 8 for HAMT. The optimizer is AdamW~\citep{loshchilov2017decoupled}. The remaining training schedules remain consistent with the baselines. Inference is performed using a greedy search approach, following a single-run setting~\citep{wang2019reinforced}. 
Note that models marked with $^*$ in table denote our re-implementation. 

\begin{table}[t]
    \setlength{\abovecaptionskip}{0.15cm}
    \setlength{\belowcaptionskip}{0cm}
    \centering
    \resizebox{0.48\textwidth}{!}{
    \begin{tabular}{c|c|c|cccc}
    \toprule
    \multirow{2}{*}{Model} & \multirow{2}{*}{\#}& \multirow{2}{*}{Alignment} & \multicolumn{4}{c}{Val-Unseen}                               \\
    \cmidrule{4-7}  &       &  &TL             & NE            & SR             & SPL            \\
    \cmidrule(lr){1-7}
    \multirow{2}{*}{DUET}  &1  &  Dual & \textbf{13.07} & \textbf{2.94} & \textbf{73.35} & \textbf{63.39} \\ \cmidrule{2-7}
                            &2  & Single & 14.42 & 2.95 & 72.88 & 61.78\\
    \bottomrule
    \end{tabular}}
    \caption{Ablation study on alignment level on the R2R dataset.}
    \label{tab:ablation_dual_level}
\end{table}
\begin{table}[]
    \setlength{\abovecaptionskip}{0.15cm}
    \setlength{\belowcaptionskip}{0cm}
    \centering
    \resizebox{0.48\textwidth}{!}{
    \begin{tabular}{l|c|cccc}
    \toprule
    \multirow{2}{*}{Model}&  \multirow{2}{*}{\#} & \multicolumn{4}{c}{Val-Unseen} \\ \cmidrule{3-6} 
                             & & TL             & NE            & SR             & SPL            \\ \cmidrule{1-6}
    DUET+DELAN              &1 & \textbf{13.07} & 2.94          & \textbf{73.35} & \textbf{63.39} \\
    \enspace w/o Instr-Land Attention & 2&15.52          & 3.05          & 72.33          & 60.42          \\
    \enspace w/o Shared Encoder  & 3&16.33          & \textbf{2.92} & 72.67          & 59.35          \\
    \bottomrule
    \end{tabular}}
    \caption{Ablation study on separation strategies for instructions and landmarks on the R2R dataset.}
    \label{tab:separation_results}
\end{table}

\subsection{Main Results}

\paragraph{Fine-grained VLN: R2R and RxR.} Table~\ref{tab:main_result_r2r} compares the performances of different agents on R2R benchmark. 
Our method outperforms baselines over all dataset splits, achieving 62.69\% SPL (+1.7\%) on the test split.
Table~\ref{tab:main_result_rxr} shows results compared with state-of-the-art methods on RxR dataset. Our method gets significant improvements (+1.1\% on SPL and +1.0\% on SR) on test split compared with the backbones. 
All above demonstrate the effectiveness and generalization of our framework. 

\paragraph{Long-horizon VLN: R4R.}
Table~\ref{tab:main_result_r4r} shows navigation results on R4R dataset. DELAN performs consistently better than the baseline HAMT (+0.9\% on SR), especially on these path fidelity related metrics (+3.3\% on CLS, +3.7\% on nDTW and +2.6\% on sDTW). This demonstrates its strong instruction following ability in long-horizon navigation scenario. DELAN can model relation between instruction and history to monitor progress, and closely follow the abundant landmarks in long instruction to navigate.
The graph-based baseline DUET naturally generates more sub-trajectories for back-and-forth exploration in the predicted path, resulting in reduced path fidelity metrics (CLS, nDTW and sDTW). However, our method can still improve almost all of its navigation metrics (especially +1.1\% on SR), which explains the universality of our framework.

\paragraph{Vision-and-Dialog Navigation: CVDN.}
The results in Table~\ref{tab:main_result_cvdn} demonstrate that our method enhances the goal progress of previous models, increasing HAMT's performance by 0.27 meters and DUET's by 0.23 meters in test environments. This highlights the effectiveness of our proposed framework in a dialog setting.

\subsection{Ablation Study}



\paragraph{Dual-level Alignment.} 
We first evaluate the necessity of employing alignment with dual levels rather than single level, as shown in Table~\ref{tab:ablation_dual_level}. Line 1 represents our DELAN framework with dual-level alignment and corresponding dual-level instruction. Line 2 depicts the use of single-level alignment, which aligns only the original instruction with concatenated visual-related components (history and observation). The results indicate that explicit alignment at dual levels significantly enhance navigation performance compared to directly aligning vision and language modalities.

\begin{table}[t]
\centering
    \setlength{\abovecaptionskip}{0.15cm}
    \setlength{\belowcaptionskip}{0cm}
    \resizebox{0.48\textwidth}{!}{
    \begin{tabular}{c|c|cccc|c|cccc}
    \toprule
    \multirow{3}{*}{Model}   
    &  \multirow{3}{*}{\#}  
    &  \multicolumn{5}{c|}{Component} 
    &  \multicolumn{4}{c}{Val-Unseen}   \\ \cmidrule{3-11}
    &  &    \multicolumn{4}{c|}{Instr-His}   
    &  Land-Ob 
    & \multirow{2}{*}{TL}  
    & \multirow{2}{*}{NE$\downarrow$}  
    & \multirow{2}{*}{SR$\uparrow$}  
    & \multirow{2}{*}{SPL$\uparrow$} \\ \cmidrule{3-7} 
    &   &  IT    &   WT &    IV  & WV   &  LO    &   &  & &    \\ \midrule
    \multirow{7}{*}{DUET}   &  1 &    \checkmark  & \checkmark   &  \checkmark    &   \checkmark &    \checkmark  & \multicolumn{1}{c}{13.07}    &   2.94   &  \textbf{73.35} &    \textbf{63.39}   \\  \cmidrule{2-11}
    &   2  & &    \checkmark  & \checkmark   &  \checkmark    &   \checkmark &    \multicolumn{1}{c}{12.99}   &  3.02  & 71.95    &   61.97  \\
    &   3  & \checkmark   &  & \checkmark   &  \checkmark    &   \checkmark &    14.58   &  2.96  & 72.84    &   60.65  \\
    &   4  & \checkmark   &  \checkmark    &   &  \checkmark    &   \checkmark &    12.84   &  3.11  & 71.56    &   62.44  \\
    &   5  & \checkmark   &  \checkmark    &   \checkmark &    &   \checkmark &    13.31   &  2.94  & 73.09    &   62.96  \\ \cmidrule{2-11}
    &   6  & \checkmark   &  \checkmark    &   \checkmark &    \checkmark  &  &    14.25   &  \textbf{2.93}  & 72.88    &   61.19  \\ \cmidrule{2-11}
    &   7  &  &  &     &   &  \checkmark   &  13.06 &    2.94    &   72.84  & 63.01  \\ \bottomrule
    \end{tabular}}
    \caption{Ablation study on different components of dual-level alignment on the R2R dataset.}
    \label{tab:ablation_contrastive}
\end{table}
\paragraph{Separation of Instruction and Landmark.} 

In the DELAN framework, we adopt a method of separating landmark words from instructions to facilitate explicit modality alignment. Table~\ref{tab:separation_results} presents performance under three approaches with increasing degrees of separation between instructions and landmarks. Line 1 illustrates the strategy employed in DELAN, allowing instructions and landmarks attending to each other using the shared text encoder. Line 2 represents a strategy where instructions and landmarks are encoded separately but with the shared text encoder. Line 3 represents a strategy utilizing two independent encoders to encode instructions and landmarks respectively, ensuring complete instruction and landmark separation before the fusion stage. The results presented in Table~\ref{tab:separation_results} highlight the superiority of the separation approach in DELAN. Since individual landmark words still hold contextual relevance, allowing them to attend to instruction words makes practical sense.


\paragraph{Components of DELAN.} We study the effectiveness of each component of DELAN over the R2R val unseen split. As shown in Table~\ref{tab:ablation_contrastive}, both instruction-history (Instr-His) and landmark-observation (Land-Ob) alignment contributes to the excellent performance of DELAN, especially on metric NE and SR.
Line 2 to 5 in Table~\ref{tab:ablation_contrastive} assess the influence of four components of the instruction-history level contrastive learning, i.e. instruction-trajectory (IT), word-trajectory (WT), instruction-viewpoint (IV), and word-viewpoint (WV). Removing the word-trajectory alignment significantly increases the trajectory length, indicating a decrease in the overall understanding of navigation progress. The elimination of instruction-viewpoint alignment degrade the success rate most, showing the importance of aligning history viewpoints with instructions. Each component has a positive effect on the performance. These results demonstrate that DELAN is not only effective but also reasonable. 

\begin{table}[t]
    \centering
    \setlength{\abovecaptionskip}{0.15cm}
    \setlength{\belowcaptionskip}{0cm}
    \resizebox{0.48\textwidth}{!}{
    \begin{tabular}{l|cccc|c}
    \toprule
    Model & Stop & Turn & Object & Room & Avg. \\
    \cmidrule(lr){1-6}
    HAMT~\citeyearpar{chen2021history} &70.62&39.61&12.01&24.96&36.80\\
    HAMT+Instr-His & 69.38 & 38.02 & 13.16 & 26.72 & 36.82\\
    HAMT+Land-Ob & 65.73 & 38.60 & 14.74 & \textbf{27.89} & 36.74\\
    HAMT+DELAN &\textbf{71.27}&\textbf{40.73}&\textbf{15.61}&25.59&\textbf{38.30}\\
    \bottomrule
    \end{tabular}}
    \caption{Quantitative analysis on the mastery of navigation skills on RxR dataset. Instr-His indicates the instruction-history alignment and Land-Ob indicates the landmark-observation alignment.}
    \label{tab:vln_behave_results}
\end{table}

\subsection{Further Analysis}
We further analyze the agent's behavior in four specific navigation skills described in~\citet{yang2023behavioral}. These skills are \textbf{Stop}, \textbf{Turn}, \textbf{Object} and \textbf{Room}, indicating the agent's ability in stopping, changing direction, object seeking, and room finding, respectively. A higher score means a stronger ability. As shown in Table~\ref{tab:vln_behave_results}, our DELAN framework significantly enhances the performance across all skills (average +1.5). We attribute the notable promotion in object seeking (+3.6) to our proposed landmark-observation level alignment, which enforces the agent to recognize landmarks appeared in panoramic observations. Meanwhile, the ability of agents to stop and change directions has also been improved, indicating that under the DELAN framework, the agent has a better judgment of instruction execution and navigation progress.

\subsection{Qualitative Results}
\begin{figure}[t]
  \setlength{\abovecaptionskip}{0.0cm}
  \centering
  \includegraphics[width=0.5\textwidth]{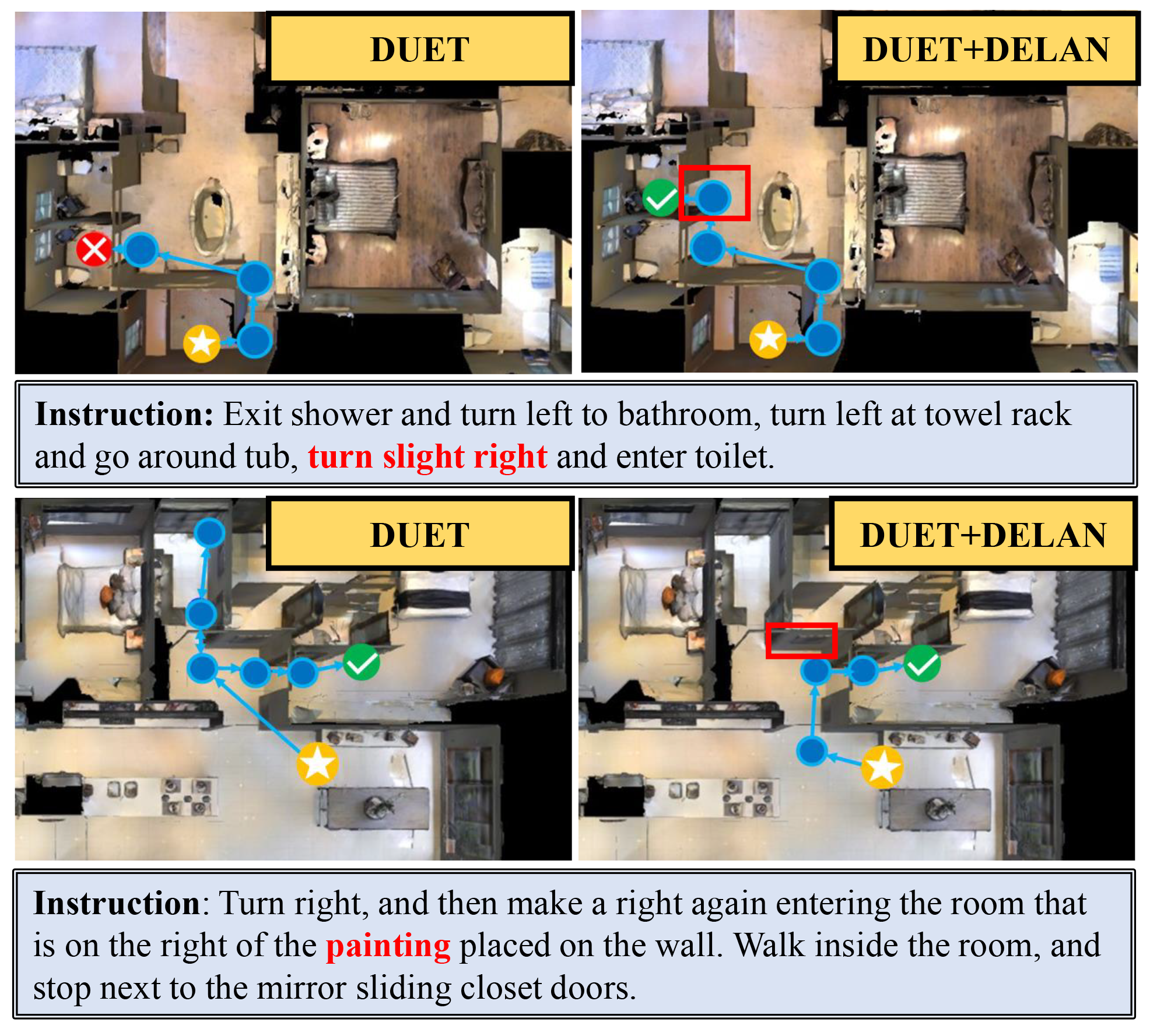}
  \caption[VLN process]{Trajectories of DUET and DUET+DELAN in R2R-unseen environments. Yellow, green, and red nodes signify the start, target, and incorrect endpoints respectively.  Red rectangles correspond the red pats of the instructions.}
  \label{fig:case}
\end{figure}

We visualize the trajectories of DUET~\citep{chen2022think} and DUET+DELAN in Figure~\ref{fig:case}. Due to the lack of the pre-fusion alignment, DUET has the unsatisfactory instruction-following capacity. For example, it fails to follow the instruction "turn slight right" or misses important landmark "painting placed on the wall", leading to either incorrect endpoints (upper) or unnecessary exploration trajectories (lower). In contrast, our proposed DELAN framework could promote the instruction comprehension and help the agent make correct decisions, resulting in a more precise and efficient navigation.

\section{Conclusion}
This paper highlights the significance of pre-fusion alignment in the VLN task and introduces a dual-level alignment (DELAN) framework, focusing on the instruction-history and landmark-observation levels. For adaptation to the dual-level alignment, we reconstruct a dual-level instruction through concatenating original instruction and landmarks. We employ different self-supervised contrastive learning strategies to realize the dual-level alignment respectively. Experiments across various VLN benchmarks demonstrate the effectiveness, generalization and universality of our approach. We hope this work can inspire further explorations on modality alignment in VLN and related fields.

\section{Acknowledgement}
This work is supported by National Natural Science Foundation of China (No.71991471, No.6217020551) and Science and Technology Commission of Shanghai Municipality Grant (No.21DZ1201402).

\section{Bibliographical References}\label{sec:reference}

\bibliographystyle{lrec-coling2024-natbib}
\bibliography{lrec-coling2024-example}


\label{lr:ref}
\bibliographystylelanguageresource{lrec-coling2024-natbib}
\bibliographylanguageresource{languageresource}

\end{document}